\documentclass[runningheads]{llncs}
\usepackage[T1]{fontenc}
\usepackage{graphicx}
\usepackage{float}
\usepackage{subcaption} 
\usepackage{algorithm}
\usepackage{algpseudocode}
\usepackage{booktabs}
\usepackage{amsmath, amssymb}
\usepackage{algpseudocode}
\usepackage{kotex}
\usepackage{booktabs}
\usepackage{bm}
\usepackage{pifont}
\usepackage{xcolor}
\usepackage{color}
\usepackage{hyperref}
\hypersetup{
    colorlinks=true,
    allcolors=black
}

\ifpdf
\input glyphtounicode
\fi
\begin{document}
\title{ConCSE: Unified Contrastive Learning and Augmentation for Code-Switched Embeddings}
\titlerunning{ConCSE for Code-Switched Embeddings}
%
\author{Jangyeong Jeon\inst{1}\orcidID{0009-0004-0124-4434} \and
Sangyeon Cho\inst{1}\orcidID{0009-0004-1072-5504} \and
Minuk Ma\inst{2}\orcidID{0000-0003-0416-8479} \and
Junyeong Kim\inst{1}\orcidID{0000-0002-7871-9627}\thanks{This work was partly supported by Institute of Information \& communications Technology Planning \& Evaluation (IITP) grant funded by the Korea government(MSIT) (No.2022-0-00184, Development and Study of AI Technologies to Inexpensively Conform to Evolving Policy on Ethics) and partly supported by Institute of Information \& communications Technology Planning \& Evaluation (IITP) grant funded by the Korea government(MSIT) (No.2021-0-01341, Artificial Intelligence Graduate School Program, Chung-Ang University}} 
\authorrunning{J. Jeon et al.}
%
\institute{Department of Artificial Intelligence, Chung-Ang University, Seoul 06974, Republic of Korea \and
Department of Computer Science, University of British Columbia, Vancouver, BC Canada V6T 1Z4 \\
\email{\{jjy6133,whtkddus98,junyeongkim\}@cau.ac.kr} \\
\email{minukma@cs.ubc.ca}}

\maketitle              
\begin{abstract}
This paper examines the Code-Switching (CS) phenomenon where two languages intertwine within a single utterance. 
There exists a noticeable need for research on the CS between English and Korean. 
We highlight that the current Equivalence Constraint (EC) theory for CS in other languages may only partially capture English-Korean CS complexities due to the intrinsic grammatical differences between the languages. 
We introduce a novel Koglish dataset tailored for English-Korean CS scenarios to mitigate such challenges. 
First, we constructed the Koglish-GLUE dataset to demonstrate the importance and need for CS datasets in various tasks. 
We found the differential outcomes of various foundation multilingual language models when trained on a monolingual versus a CS dataset. 
Motivated by this, we hypothesized that SimCSE, which has shown strengths in monolingual sentence embedding, would have limitations in CS scenarios. 
We construct a novel Koglish-NLI (Natural Language Inference) dataset using a CS augmentation-based approach to verify this. 
From this CS-augmented dataset Koglish-NLI, we propose a unified contrastive learning and augmentation method for code-switched embeddings, ConCSE, highlighting the semantics of CS sentences. 
Experimental results validate the proposed ConCSE with an average performance enhancement of 1.77\% on the Koglish-STS(Semantic Textual Similarity) tasks. 
\footnote{Source code available at \url{https://github.com/jjy961228/ConCSE}}

\keywords{Natural Language Processing \and Contrastive Learning \and Code-Switching \and Bilingual \and Koglish dataset }
\end{abstract}

\section{Introduction}
Code-switching (CS) refers to the phenomenon of two languages intermixed within a single sentence~\cite{auer2013code,POPLACK+1980+581+618}. Such occurrences are frequently observed in multicultural countries, social media, and online platforms~\cite{baker2011foundations,POPLACK+1980+581+618,ahn2017language}.
According to recent findings, despite the growing interest in CS, there remains a dearth of related studies~\cite{sebba2012language}.
Especially in countries where English is not the dominant language, the phenomenon of CS between English and the native language is particularly prominent ~\cite{POPLACK+1980+581+618,baker2011foundations,park1990korean,miwa1985intrasentential,amazouz2017addressing,Eunsun2021korean}.
For example, the English sentence ``\textit{The movie} was very dull'' can be represented as ``\textit{영화} was very dull.'' for English-Korean and ``\textit{la película} was very dull.'' for English-Spanish.

Past research introduced the Equivalence Constraint (EC) theory as a condition for the occurrence of CS~\cite{POPLACK+1980+581+618}, prompting attempts to construct CS datasets based on The EC theory~\cite{pratapa2018language,srinivasan2020code}.
The EC theory posits that switches between languages in a code-switched discourse tend to happen at points where the grammatical structures of the involved languages match. 
According to the EC theory, such alignment in grammatical structures demonstrates that code-switching adheres to systematic linguistic constraints.
This foundational concept has been central in many CS studies, particularly language pairs like English-Spanish and English-Chinese~\cite{pratapa2018language,POPLACK+1980+581+618,winata2019code,hu2020data,winata2018code}.
However, studies on CS between English and Korean show that this assumption is not always met~\cite{Eunsun2021korean}.
For English-Korean CS, there is a potential limitation that EC Theory does not satisfy due to the grammatical difference between the two languages.
For instance, the grammatical differences between English and Korean primarily manifest in word order and case marking. English predominantly follows an SVO (Subject-Verb-Object) word order, and this sequence largely determines the meaning of a sentence. 
In contrast, Korean offers greater flexibility in the positioning of subjects and objects, thanks in large part to its distinctive case markers like “이[i]/가[ga]” (nominative), “을[eul]/를[leul]” (accusative), and “에게[ege]” (dative). 
Crucially, altering the word order in English can significantly change the meaning of a sentence, whereas, in Korean, where the language's case markers are well developed, position shifts within sentence components are accessible~\cite{lehmann1973structural,Eunsun2021korean}.

This paper introduces a novel Koglish dataset and proposes a new approach to constructing CS datasets, considering the inherent complexity of CS.
The Koglish dataset includes Koglish-GLUE, Koglish-NLI, and Koglish-STS datasets. 
In particular, we propose to apply constituency parsing~\cite{joshi2018extending} to construct the Koglish dataset to obtain parse trees and transform English sentences into CS sentences following the approach proposed in Sect.~\ref{sec3.2}.
To construct the Koglish dataset, We utilize GLUE benchmark~\cite{wang2018glue},
Semantic Textual Similarity (STS)~\cite{agirre2012semeval,agirre2013sem,agirre2014semeval,agirre2015semeval,agirre2016semeval,marellisick},
The Stanford Natural Language Inference Corpus 
 (SNLI)~\cite{bowman2015large}, and The Multi-Genre Natural Language Inference Corpus (MNLI)~\cite{williams2017broad}. 
To better understand the need for code-switching (CS) datasets, we posited the following hypothesis:
There will be a noticeable difference in performance between training with a monolingual dataset and then testing on a CS dataset (EN2CS) versus conducting both training and testing with a CS dataset (CS2CS). This significant disparity underscores the importance of using our CS dataset, Koglish, in CS scenarios.
To our knowledge, this is the first presentation of Koglish datasets suitable for English-Korean and Korean-English scenarios.

Determining semantic relationships between sentences is a critical challenge in natural language processing. 
Recently, contrastive learning drew significant attention in natural language processing~\cite{gao2021simcse,wu2021esimcse,chuang2022diffcse}, where the model learns to distinguish between pairs of similar and dissimilar samples. 
For example, SimCSE~\cite{gao2021simcse} proposed to convert the sentence pairs of (premise, hypothesis) in the Natural Language Inference (NLI) dataset~\cite{bowman2015large,williams2017broad} into the triplets of (premise, entailment, contradiction) to provide extra signals for contrastive learning.
However, the study of contrastive learning under code-switched sentences has been largely yet to be underexplored.
To address this issue, we propose a unified contrastive learning and data augmentation method dubbed ConCSE to model the code-switched sentences explicitly.
For each sentence triplet of (premise, entailment, contradiction), we generate a triplet of code-switched sentences (CS-premise, CS-entailment, CS-contradiction) via CS-augmentation in Sect.~\ref{sec3.2} using a constituency parser.
Then it considers the relationships between the six sentences to define three novel loss functions: (1) Cross Contrastive Loss $(\mathcal{L}_{CS}^{Con})$, (2) Cross Triplet Loss $(\mathcal{L}_{CS}^{Tri})$, and (3) Align Negative Loss $(\mathcal{L}^{Sim}_{neg})$, providing richer supervision compared to plain SimCSE. 
For example, the sentence pairs of (premise, CS-premise) are considered positive, while those of (CS-premise, contradiction) are considered negative. 
As a validation, we compared the performance of four baseline multilingual models across seven NLP tasks included in Koglish-STS. 
The baseline multilingual models struggle to perform on the code-switched scenarios, suggesting the intricacy and effectiveness of the Koglish dataset.
The experiments on the ConCSE method on the Koglish-STS dataset show consistent performance improvements over SimCSE across seven semantic textual similarity (STS) tasks included in Koglish-STS.

Our contributions can be summarized as follows:
\begin{itemize}
\item We introduce the first dataset referred to as Koglish which is suitable for English-Korean and Korean-English CS scenarios including Koglish-GLUE\footnotemark[4], Koglish-STS\footnotemark[5]\footnotemark[6], and Koglish-NLI\footnotemark[7].
\item We demonstrate the necessity of the Koglish dataset through various experiments.
\item We propose an effective sentence representation learning method that considers the CS sentences through a specialized CS-focused augmentation technique.
\end{itemize}

\footnotetext[4]{\url{https://huggingface.co/datasets/Jangyeong/Koglish\_GLUE}}
\footnotetext[5]{\url{https://huggingface.co/datasets/Jangyeong/Koglish\_STS}}
\footnotetext[6]{\url{https://huggingface.co/datasets/Jangyeong/Koglish\_SICK}}
\footnotetext[7]{\url{https://huggingface.co/datasets/Jangyeong/Koglish\_NLI}}

\section{Related Work}
\subsection{Theoretical Foundations of Code-Switching}
In previous research, the conditions for the occurrence of Code-Switching (CS) and Code-Mixing (CM) were proposed as the Equivalence Constraint (EC) theory, Matrix Language Framework (MLF), and Functional Head Constraint.
Notably, when the EC Theory criteria are met, studies have constructed CS and CM datasets using a Constituency parser~\cite{srinivasan2020code,pratapa2018language}.
This approach has found application in English-Chinese Code-Switching studies as well~\cite{winata2019code,pratapa2018language}.
However, investigations into English-Korean CS have demonstrated that most instances do not conform to the EC theory, indicating its unsuitability for English-Korean CS scenarios~\cite{park1993constraints,park1990korean,Eunsun2021korean}.
The research highlights that in English-Korean and Korean-English code-switching, nouns or noun phrases often serve as the Embedded Language (EL), with their usage being notably prevalent, accounting for 74.6\% and 61\% respectively~\cite{park1993constraints,Eunsun2021korean}. 
These prior empirical results showed the importance of selecting nouns or noun phrases as EL in constructing an English-Korean CS dataset.
Pursuing this approach, our study uses a pre-trained Constituency parser~\cite{joshi2018extending} to identify and extract nouns or noun phrases.

\subsection{Representation Learning}
\subsubsection{Deep Metric Learning}
Deep Metric Learning was formulated to decipher the dynamics of embedding spaces~\cite{chopra2005learning,hadsell2006dimensionality,weinberger2009distance}. 
Among its diverse strategies, triplet loss stands out~\cite{hoffer2015deep}. It emphasizes the interrelationships and distances of samples within the embedding space, aiming to cluster similar samples and distance dissimilar ones closely.
A pivotal element in this approach is the 'margin,' a hyperparameter designed to ensure a defined distance between the anchor-positive and anchor-negative pairs~\cite{schroff2015facenet}. 
This paper utilizes triplet loss as an auxiliary loss to bolster the model’s stability.

\subsubsection{Contrastive Learning}
In fields like natural language processing~\cite{gao2021simcse,chuang2022diffcse,wu2021esimcse} and computer vision~\cite{chen2020simple,kuang2021video}, the core aim is to enhance representations by discerning between positive and negative samples. 
Contrastive learning, which builds upon the foundations of deep metric learning, offers refined techniques for achieving superior representations. 
A notable advancement is the introduction of data augmentation to enrich training datasets. 
While random cropping and image rotation succeed in computer vision~\cite{chen2020simple,yan2021consert}, their adaptation to natural language processing poses challenges. 
Nevertheless, strategies reconstructing NLI datasets for contrastive learning have been proposed to bridge this gap~\cite{gao2021simcse,chuang2022diffcse}.
In particular, in the strategy of reconstructing NLI datasets~\cite{gao2021simcse,chuang2022diffcse}, first, all datasets labeled as neutral are excluded, and only datasets labeled as entailment for two sentences (premise, hypothesis) are extracted.
In this case, the premise and hypothesis are defined as a positive pair, and the hypothesis is defined as an entailment sentence. 
Second, extract hypothesis sentences where the hypothesis is labeled as a contradiction for the same sentence as the premise used in the first step. In this case, the premise and hypothesis are defined as a negative pair, and the hypothesis is defined as a contradiction sentence.
The NLI dataset was redefined as premise, entailment, and contradiction sentences and used for training SimCSE.
Yet, the proposed strategies of SimCSE are limited to monolingual datasets. To address this limitation, our study presents a novel method: augmenting a resource-rich English dataset with a CS dataset in a supervised setting.

\vspace{-0.2cm}
\section{Proposed Dataset: Koglish} \label{sec3 : Dataset}

\begin{figure}[!tp]
\centering
\resizebox{\columnwidth}{!}{
\includegraphics{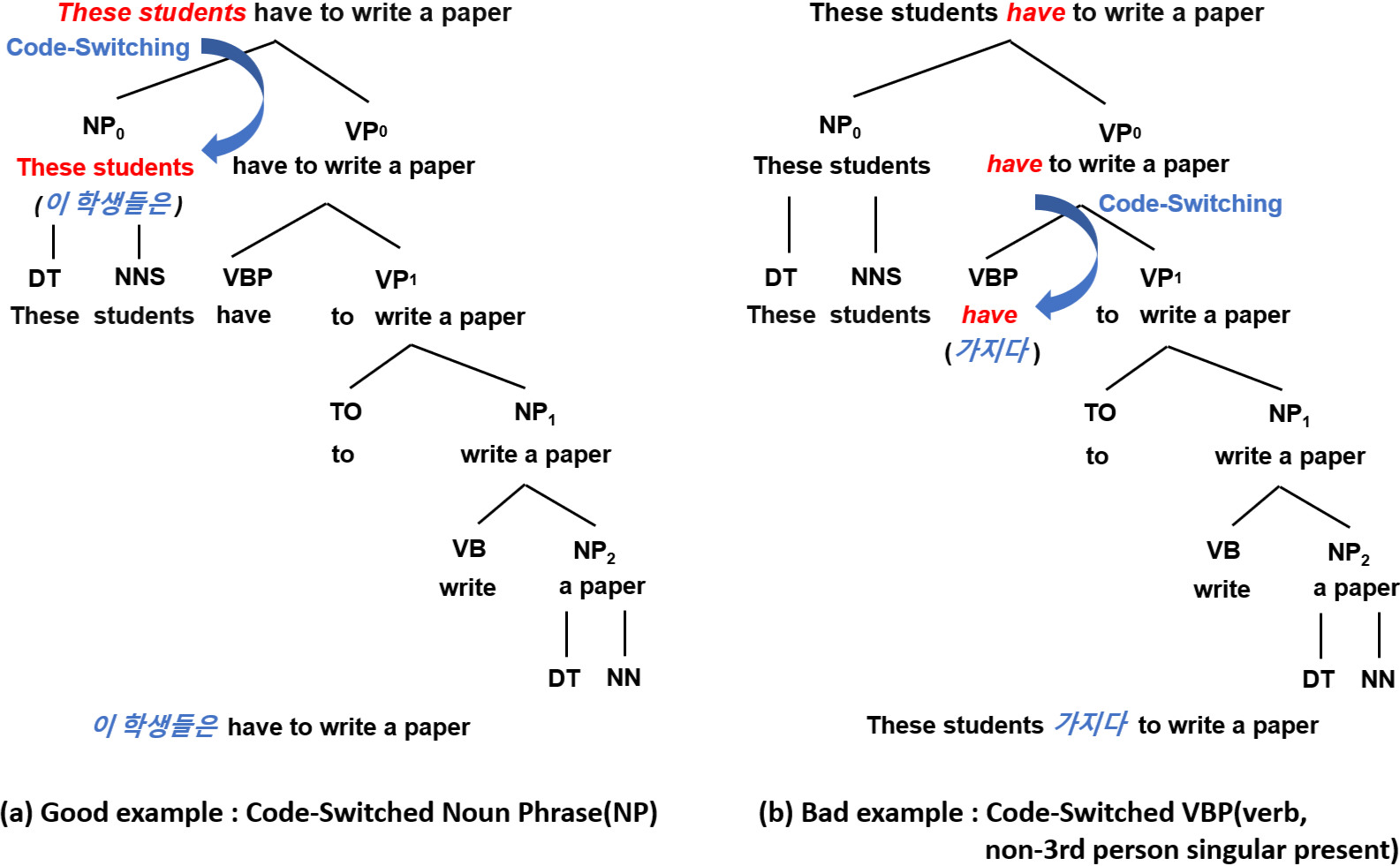}
}
\caption{Schematic of the parse tree, based on constituency parsing, to convert a monolingual sentence into an English-Korean code-switched sentence.}
\label{fig.1: parse tree}
\end{figure}

This section elaborates on English-Korean and Korean-English code-switching sentences and our specialized Koglish dataset construction and CS augmentation strategies.  
A summary of the constructed dataset is provided in Table~\ref{table.1}.

\subsection{Code-switching Patterns and Dataset Construction}
According to a study by~\cite{park1990korean}, Code-Switching (CS) between English and Korean does not adhere to the guidelines established by the EC Theory~\cite{POPLACK+1980+581+618} and the Matrix Language Frame (MLF) Model~\cite{myers1990intersections}.
This is due to the fact that the grammatical units (e.g., phrase, adjective phrase, verb phrase) converted in CS are language-specific.
Consequently, when constructing CS datasets, it is imperative to use strategies tailored to each respective language~\cite{park1990korean,POPLACK+1980+581+618,miwa1985intrasentential,park1993constraints,amazouz2017addressing}.
Historical analyses indicate that in Korean-English CS, nouns and noun phrases constitute 74.6\% of code-switched instances~\cite{park1993constraints}. 
English-Korean exhibits a similar trend, with nouns representing 61\% of code-switched ~\cite{Eunsun2021korean}. 
As shown in Fig.\ref{fig.1: parse tree}-(a), code-switching the noun phrase maintains the sentence's integrity, mirroring the structure of the original. 
In contrast, code-switching VBP(Verb, non-3rd person singular present) as shown in Fig.\ref{fig.1: parse tree}-(b), produces a sentence that is awkwardly constructed.
Japanese, sharing syntactic similarities with Korean, also has a high noun switching rate at 68.8\%~\cite{miwa1985intrasentential}. 
This structural congruence suggests the potential for applying our CS dataset construction strategy to other languages with grammatical structures akin to Korean's~\cite{Eunsun2021korean}.
In contrast, Spanish-English code-switching contains a significantly lower noun switch rate, sometimes reaching lower than 20\%~\cite{POPLACK+1980+581+618}.
Given these patterns, we primarily focused on switching nouns or noun phrases when constructing English-Korean and Korean-English CS datasets.
Additionally, due to the distinction between Matrix Language (ML) and Embedded Language (EL) is not explicit in English-Korean code-switching~\cite{myers1990intersections}, the dominant use of nouns and noun phrases in both English-Korean and Korean-English code-switching endorses the suitability of our proposed dataset strategy for both scenarios.

\begin{table}[!t] 
\begin{center}
\setlength\tabcolsep{12pt} 
\begin{tabular}{l|c c c|c} %
  \multicolumn{5}{c}{\textbf{GLUE Benchmark}} \\ 
  \textbf{Task} & \textbf{Train} & \textbf{Dev} & \textbf{Test} & \textbf{Total}\\
  \hline
  QNLI &  61,764  &  15,441  &  19,303 & \textbf{96,508} \\
  \hline
  SST-2 &  26,552 &  6,632 &  8,289  & \textbf{41,473} \\
  \hline
  COLA &  4,341  &  1,087 &  1,358  & \textbf{6,759} \\
  \hline
  STS-B &  4,269  &  1,068  &  1,335 & \textbf{6,672} \\
  \hline
  MRPC &  3,610 &  904 &  1,129 & \textbf{5,643} \\
  \hline
  RTE &  1,642  &  412  &  514  & \textbf{2,568} \\
  \hline
  \multicolumn{5}{c}{\textbf{Semantic Textual Similarity(STS)}} \\
  
  \textbf{Task} & \textbf{Train} & \textbf{Dev} & \textbf{Test} & \textbf{Total}\\
  \hline
  STS-B &  \quad-  &  3,334  &  3,334  & \textbf{6,668} \\
  \hline
  STS12 &  \quad-  &  2,142  & 2,142 & \textbf{4,286} \\
  \hline
  STS13 & \quad- & 622 & 622 & \textbf{1,244} \\
  \hline
  STS14 & \quad- & 1,561 & 1,561 & \textbf{3,112} \\
  \hline
  STS15 & \quad- & 1,351 & 1,351 & \textbf{2,702} \\
  \hline
  STS16 & \quad- & 496 & 496 & \textbf{992} \\
  \hline
  SICK-R & \quad- & 4,767 & 4,767 & \textbf{9,534} \\
  \hline
  \multicolumn{5}{c}{\textbf{Natural Language Inference(NLI)}} \\
  
  \textbf{Task} & \textbf{Train} & \textbf{Dev} & \textbf{Test} & \textbf{Total}\\
  \hline
  NLI & 218,255 & \quad- & \quad- & \textbf{218,255} \\
\end{tabular}
\caption{Summary of Koglish Datasets.}
\label{table.1}
\end{center}
\end{table}

\subsection{Constructing Koglish Dataset}
\begin{figure*}[t]
        \includegraphics[width=12.5cm,height=10cm]{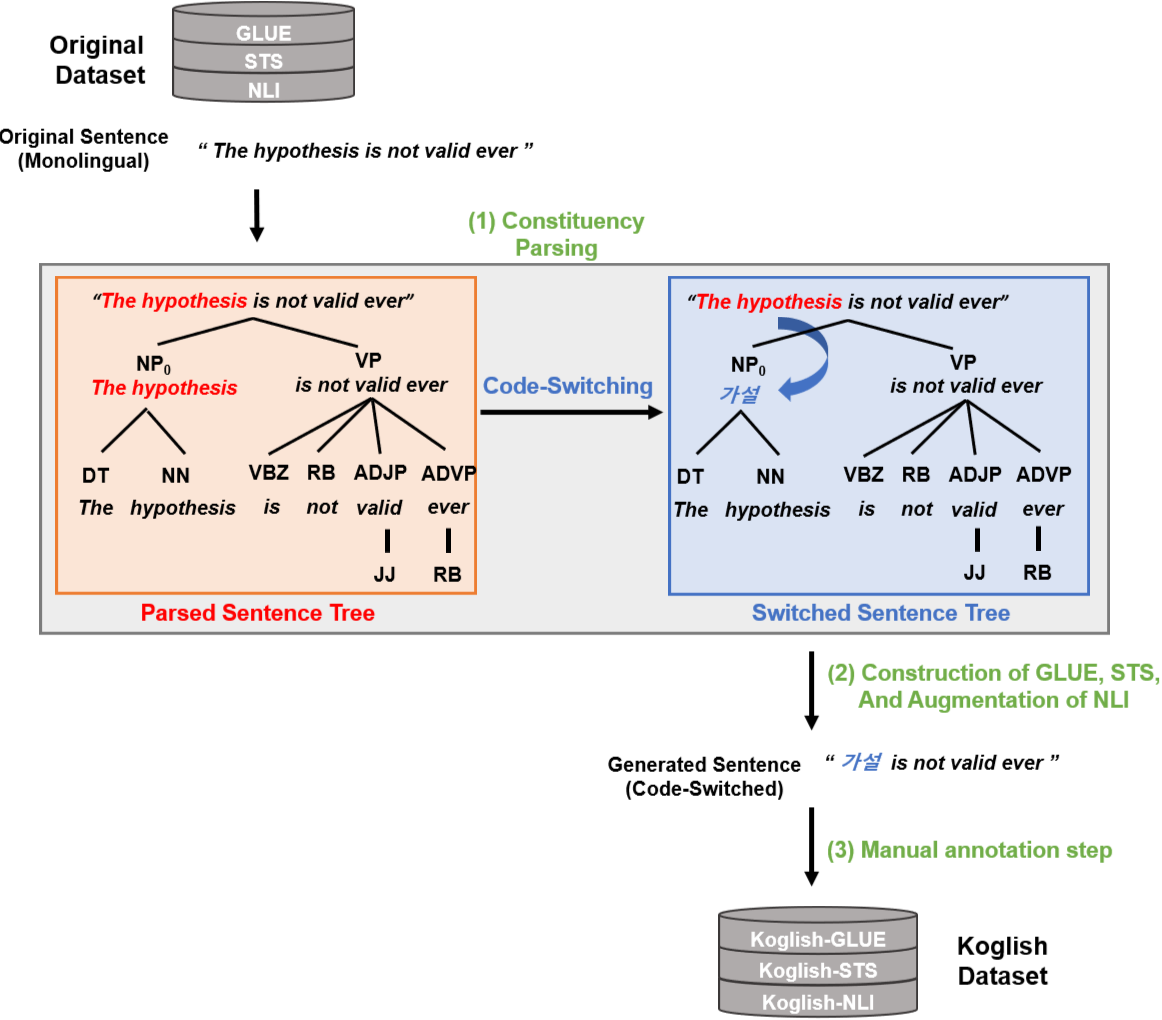} 
        \caption{Systematic Approach to Constructing and Augmenting the Koglish Dataset. Constituency parser extracts nouns or noun phrases(NP) using the Google Translate API. In this case, GLUE and STS datasets are generated as CS datasets, and NLI datasets are CS-augmented.}
        \label{fig.2}
\end{figure*}
\label{sec3.2}
This section details constructing and augmenting the proposed CS dataset, Koglish. The overall process is shown in Fig.~\ref{fig.2}. 
\begin{enumerate}  
    \item We constructed a parse tree using a top-down constituency parsing approach~\cite{joshi2018extending}. During this process, we selectively extracted the $NP$ nodes, ensuring the inclusion of nouns and noun phrases (see Fig.~\ref{fig.2}-(1)). In some data, entire sentences were constructed solely from the $NP$ structure. When such sentences underwent the translation process, they resulted in monolingual sentences, negating the goal of CS. 
    Therefore, we excluded these particular entries. Additionally, if the $NP_0$ node contained only pronouns (e.g., It, That, This), it led to mistranslation issues. 
    To address this, we extracted the $NP$ node from the subsequent $NP_1$ node and applied the top-down approach to the leaf nodes. 
    If the data did not align with our criteria when it reached the leaf node, we considered it inappropriate for the CS dataset and subsequently excluded it.
    For example, GLUE’s COLA task data excluded 29.1\% of the entire data.
    
    \item Generate CS sentences from the Switched Sentence Tree of Fig.~\ref{fig.2}-(1) as shown in Fig.~\ref{fig.2}-(2). 
    The first is the GLUE~\cite{wang2018glue} and STS dataset~\cite{agirre2012semeval,agirre2013sem,agirre2014semeval,agirre2015semeval,agirre2016semeval,marellisick} , and the second is the NLI~\cite{bowman2015large,williams2017broad} dataset. 
    As an example of the first, GLUE and STS take monolingual sentences as input and generate a CS sentence if it satisfies the abovementioned conditions (in step 1).
    The second example is the NLI dataset, which receives triplets of monolingual sentences (e.g., premise, entailment, and contradiction) as input.
    If the above conditions (in step 1) are satisfied for the triplet of monolingual sentences, it generates CS-premise, CS-entailment, and CS-contradiction. 
    In the following Sect.~\ref{section: ConCSE}, the three sentences (premise, entailment, and contradiction) of the Koglish-NLI dataset and CS-Augmented sentences (CS-premise, CS-entailment, and CS-contradiction) are integrated, and used for learning ConCSE, so in this paper, we assume that only NLI is CS-Augmented sentences.
        
    \item To ensure reliability and accuracy, we performed critical manual annotations on the generated Koglish datasets. This process involved bilingual experts proficient in both Korean and English. We employed four annotators, each tasked with evaluating the contextual accuracy of the Code-Switching sentences in the dataset. Following their assessments, the four annotators produced each dataset through a meticulous cross-validation process, rigorously examining each other’s evaluations(see Fig.~\ref{fig.2}-(3)). 
    Finally, we split each dataset. 
    The Koglish-GLUE was divided into train, development, and test sets in the ratios of 0.64, 0.16, and 0.20, respectively, to formulate the Koglish-GLUE dataset.
    Since the Koglish-STS dataset is only used to evaluate ConCSE in Sect.~\ref{Experiments : ConCSE}, we constructed the Koglish-STS dataset by splitting the development and test sets equally (0.5 ratios each).
    We constructed Koglish-NLI without any segmentation since the Koglish-NLI dataset is only used for training.
\end{enumerate}

\section{Proposed Method: ConCSE}
\label{section: ConCSE}
    \begin{figure}[!tp] 
    \centering
        \includegraphics[width=11.5cm, height=5.5cm]{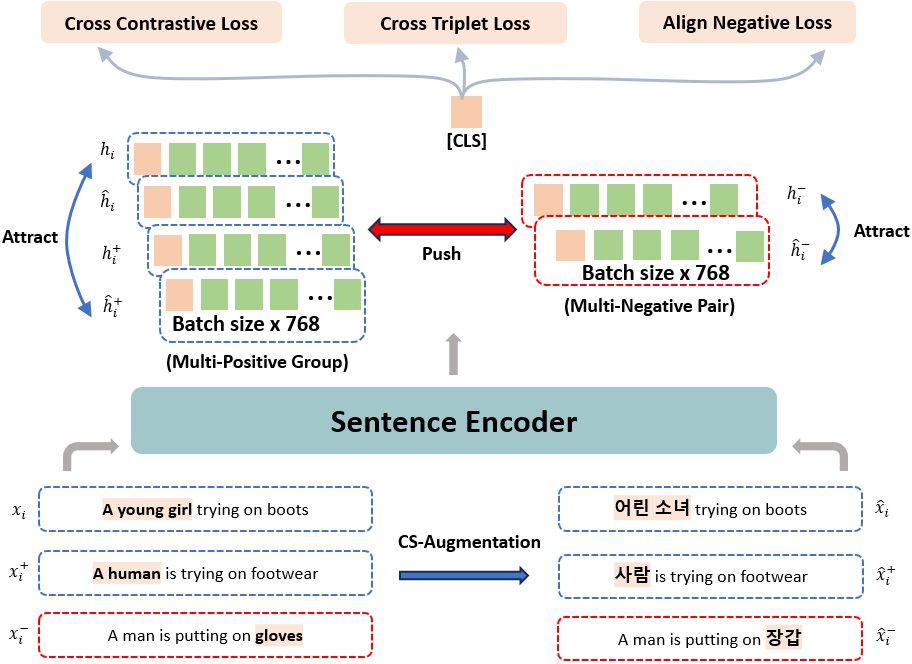} 
    \caption{Overview of ConCSE. A mini-batch contains both $\mathcal{D}_{en} = \{x_i, x_{i}^{+}, x_{i}^{-}\}_{i=1}^{m}$ and CS-augmented $\mathcal{D}_{cs} = \{\hat{x}_i, \hat{x}_{i}^{+}, \hat{x}_{i}^{-}\}_{i=1}^{m}$, and its hidden representations are $H = \{h_{i},h_{i}^{+},h_{i}^{-}\}_{i=1}^{N}$ and $\hat{H} = \{\hat{h}_{i},\hat{h}_{i}^{+},\hat{h}_{i}^{-}\}_{i=1}^{N}$. They are processed by the sentence encoder $\mathcal{M}_{\phi}$, producing ``[CLS]'' as the final sentence representation. 
    The ``[CLS]'' of the multi-positive group, comprising monolingual sentences ($h_i, h_{i}^{+}$) and CS sentences ($\hat{h}_i, \hat{h}_{i}^{+}$), should be attracted to each other. 
    Similarly, the ``[CLS]'' of the multi-negative pair, comprising a monolingual sentence ($h_{i}^{-}$) and CS sentence ($\hat{h}_{i}^{-}$), should also be attracted to each other. Moreover, multi-positive groups and multi-negative pairs should push each other.
    }
    \label{fig.model}
    \vspace{-0.3cm}
    \end{figure}
This paper aims to train universal sentence embeddings in Code-Switching (CS) contexts. 
As detailed in step 2 of Sect.~\ref{sec3.2}, we use the monolingual datasets $\mathcal{D}_{en} = \{x_i, x_{i}^{+}, x_{i}^{-}\}_{i=1}^{m}$ and the augmented CS datasets $\mathcal{D}_{cs} = \{\hat{x}_i, \hat{x}_{i}^{+}, \hat{x}_{i}^{-}\}_{i=1}^{m}$ to fine-tune a pre-trained multilingual sentence encoder $\mathcal{M}_{\phi}$, such as mBERT~\cite{devlin2018bert} or XLM-R~\cite{conneau2020unsupervised}, to adapt to the CS scenario.

The notation for the comprehensive loss function used is:
\begin{equation}
\begin{aligned}
\mathcal{L}_{total} = \mathcal{L}_{CS}^{Con} + \lambda\mathcal{L}_{CS}^{Tri} + \mathcal{L}^{Sim}_{neg}
\end{aligned}
\end{equation}
where $\lambda$ signifies the weight factor assigned to the triplet loss. Detailed explanations of $\mathcal{L}_{CS}^{Con}$, $\mathcal{L}_{CS}^{Tri}$, and $\mathcal{L}^{Sim}_{neg}$ can be found in Sect.~\ref{subsec: Cross Contrastive Loss},~\ref{subsec: Cross Triplet Loss}, and~\ref{subsec: Align Negative Loss}, respectively.
An overview of ConCSE is shown in Fig.~\ref{fig.model}.

\subsection{Cross Contrastive Loss}
\label{subsec: Cross Contrastive Loss}

We train $\mathcal{M}{\phi}$ with Cross Contrastive Loss ($\mathcal{L}_{CS}^{Con}$) on monolingual and CS sentences. 
The hidden state of ``[CLS]'' for \(\mathcal{D}_{en}\) within \(\mathcal{M}_{\phi}\) is defined as: 
\begin{equation}
\begin{aligned}
H = \{h_{i},h_{i}^{+},h_{i}^{-}\}_{i=1}^{N}
\end{aligned}
\end{equation}
For \(\mathcal{D}_{cs}\) within \(\mathcal{M}_{\phi}\), it is defined as :  
\begin{equation}
\begin{aligned}
\hat{H} = \{\hat{h}_{i},\hat{h}_{i}^{+},\hat{h}_{i}^{-}\}_{i=1}^{N}
\end{aligned}
\end{equation}

where \(N\) is the batch size.
This paper extends contrastive loss to include six combinations, facilitating cross-training on 
$\mathcal{D}_{en}$ and $\mathcal{D}_{cs}$ : 
\begin{equation}
\begin{aligned}
& \mathcal{H}^1= \{h_{i}, h_{i}^{+},h_{i}^{-}\}_{i=1}^{N}  ,  \mathcal{H}^2=\{\hat{h}_i, \hat{h}_{i}^{+},\hat{h}_{i}^{-}\}_{i=1}^{N}, \\
& \mathcal{H}^3=\{h_{i}, h_{i}^{+},\hat{h}_{i}^{-}\}_{i=1}^{N}  ,  \mathcal{H}^4=\{\hat{h}_i, \hat{h}_{i}^{+},h_{i}^{-}\}_{i=1}^{N}, \\
& \mathcal{H}^5=\{h_{i}, \hat{h}_{i},h_{i}^{-}\}_{i=1}^{N}  ,  \mathcal{H}^6=\{h_{i}^{+}, \hat{h}_{i}^{+},\hat{h}_{i}^{-}\}_{i=1}^{N} \\
\end{aligned}
\label{eq: combination}
\end{equation}

For instance, the loss function $\mathcal{L}_{\mathcal{H}^{3}}^{con}$, which cross-trains on $\mathcal{D}_{en}$ and $\mathcal{D}_{cs}$, can be denoted as:
\vspace{-0.1cm}
\begin{equation}
\resizebox{.70\hsize}{!}{%
    \ensuremath{ 
    \begin{aligned}
        \mathcal{L}_{\mathcal{H}^{3}}^{Con}= 
        \sum_{i=1}^{N}- \log \frac{e^{\mathrm{sim}(\mathbf{h}_i,\mathbf{h}_i^+ )/ \tau }}{\sum_{j=1}^N\left(e^{\mathrm{sim}(\mathbf{h}_i,\mathbf{h}_j^+)/\tau}+e^{\mathrm{sim}(\mathbf{h}_i,\mathbf{\hat{h}}_j^-)/ \tau}\right)}
    \end{aligned}
    }
}
\label{eq:simcse}
\end{equation}
where, $\mathrm{sim}(\cdot,\cdot)$ is the cosine similarity function.

By integrating from Equation~\ref{eq: combination} and~\ref{eq:simcse}, the Cross Contrast Loss $\mathcal{L}_{CS}^{Con}$ is calculated as follows: 
\begin{equation}
    \label{eq:Cross contarstive loss}
    \resizebox{.24\hsize}{!}{
    \ensuremath{
        \begin{aligned}
            \mathcal{L}_{CS}^{Con} = 
            \sum_{k=1}^{6} \mathcal{L}_{\mathcal{H}^{k}}^{Con}
        \end{aligned}
    }}
\end{equation}

\vspace{-0.6cm}
\subsection{Cross Triplet Loss}
\label{subsec: Cross Triplet Loss}
Following the proposed Cross Contrastive Loss ($\mathcal{L}_{CS}^{Con}$), the triplet loss~\cite{schroff2015facenet} is introduced to adjust the distance between the anchor and positive and the distance between the anchor and negative by a margin ($\alpha$). 
Triplet loss can be extended to six combinations, as in Equation~\ref{eq: combination}, to allow cross-training on $\mathcal{D}_{en}$ and $\mathcal{D}_{cs}$. An example for $\mathcal{L}_{\mathcal{H}^{3}}^{Tri}$ is defined as:
\vspace{-0.1cm}
\begin{equation}
\label{eq : triplet loss}
 \mathcal{L}_{\mathcal{H}^{3}}^{Tri} = \sum_{i=1}^{N} \max(0, \|{h_{i}} - {h_{i}^{+}}\|_2^2 - \|{h_{i}} - {\hat{h}_{i}^{-}}\|_2^2 + \alpha)
\end{equation}
where $N$ is the batch size. To this end, $\mathcal{L}_{CS}^{Tri}$, derived from Equation~\ref{eq: combination} and~\ref{eq : triplet loss}, is defined as:

\vspace{-0.4cm}
\begin{equation}
    \label{eq: Cross Triplet loss}
    \mathcal{L}_{CS}^{Tri} = 
    \sum_{k=1}^{6} \mathcal{L}_{\mathcal{H}^{k}}^{Tri}
\end{equation}

\subsection{Align Negative Loss}
\label{subsec: Align Negative Loss}
The negative samples from \(\mathcal{D}_{en}\) and \(\mathcal{D}_{cs}\) should share the same meaning, implying that they should be in a positive relationship with each other. We define the loss function $\mathcal{L}^{sim}_{neg}$ to encode this relationship into the \(\mathcal{M}_{\phi}\):
\begin{equation}
   \mathcal{L}^{sim}_{neg} = 
    \sum_{i=1}^{N} CE(sim(h_{i}^{-},\hat{h}_{i}^{-})) 
\end{equation}
where $CE(\cdot)$ denotes cross-entropy loss, $sim(\cdot,\cdot)$ is the cosine similarity function, and \(N\) is the batch size.

\section{Experiments}
\subsection{Experiments on Koglish: The Role of Koglish in Code-Switching Scenario}
\label{Experiments: CS Dataset}

\subsubsection{Setup}
In this experiment, we utilize our Koglish-GLUE dataset. 
Considering the MRPC task as an example, which determines if a pair of sentences in the Koglish-GLUE dataset are semantically equivalent: this task comprises the original English sentences, namely sentence0 and sentence1 from GLUE, as well as the Code-Switched (CS) versions, CS-sentence0 and CS-sentence1.
For example, in the EN2CS scenario, we perform training and evaluation using only the monolingual English dataset sentence0 and sentence1. 
In the EN2CS scenario, sentence0 and sentence1 serve as the training data, while CS-sentence0 and CS-sentence1 are utilized for evaluation. 
Detailed information regarding the data used in the experiments is provided in Table~\ref{table.1}.
The evaluation metrics for each experiment align with those adopted in BERT~\cite{devlin2018bert}. Specifically, the MRPC uses the F1-score, STS-B uses Spearman’s correlation, and the remaining tasks rely on accuracy for performance measurement.

The central hypothesis of this experiment is twofold. First, if the proposed CS dataset construction method is valid, the performance difference between English to English(EN2EN) and Code-Switching to Code-Switching(CS2CS) should be insignificant.  
Second, if CS2CS’s performance is better than EN2CS’s, this suggests the need for fine-tuning using proposed CS datasets in a CS scenario.

\begin{table*}[!tp] 
\begin{center}
\setlength{\tabcolsep}{3pt}
\renewcommand{\arraystretch}{1.8}
\resizebox{\columnwidth}{!}{
\begin{tabular}{l|c c c c c c|c} 
  \multicolumn{8}{c}{\textbf{English to English(EN2EN)}} \\
  \textbf{Model} & \textbf{COLA} & \textbf{SST-2} & \textbf{MRPC} & \textbf{RTE} & \textbf{STS-B} & \textbf{QNLI} & \textbf{Avg.} \\ 
  \hline
  mBERT\textsubscript{base}& 74.66$\pm$1.52 & 92.62$\pm$0.47 & 87.23$\pm$2.24 & 66.28$\pm$1.11 & 87.18$\pm$0.69 & 88.48$\pm$0.26 & 82.74 \\ 
  XLM-R\textsubscript{base}& 72.37$\pm$0.59 & 93.85$\pm$0.32 & 88.38$\pm$0.69 & 60.27$\pm$1.97 & 87.23$\pm$0.46 & 87.87$\pm$0.19 & 81.66 \\ 
  XLM-R\textsubscript{large}& 79.90$\pm$4.03 & 95.10$\pm$0.14 & 89.53$\pm$0.80 & 65.34$\pm$4.14 & 90.51$\pm$0.28 &91.82$\pm$0.14 & 85.37 \\ 
  mBART\textsubscript{large}& 78.94$\pm$0.40 & 94.29$\pm$0.09 & 89.32$\pm$0.31 & 69.40$\pm$1.10 & 89.24$\pm$0.32 & 90.83$\pm$0.09 & 85.34 \\ 
  
  \multicolumn{8}{c}{\textbf{English to Code-Switching(EN2CS)}} \\ 
  \textbf{Model} & \textbf{COLA} & \textbf{SST-2} & \textbf{MRPC} & \textbf{RTE} & \textbf{STS-B} & \textbf{QNLI} & \textbf{Avg.} \\
  \hline
  mBERT\textsubscript{base}& 68.59$\pm$3.86 & 81.48$\pm$2.04 & 79.18$\pm$2.43 & 56.10$\pm$1.86 & 74.42$\pm$1.51 & 77.75$\pm$0.6 & 72.92 \\ 
  XLM-R\textsubscript{base}& \textbf{72.20$\pm$0.27} & 88.55$\pm$0.46 & 83.65$\pm$1.84 & \textbf{54.58$\pm$1.18} & 78.84$\pm$1.04 & 79.44$\pm$0.4 & 76.21 \\ 
  XLM-R\textsubscript{large}&\textbf{74.61$\pm$1.98} & 91.78$\pm$0.13 & 87.98$\pm$0.49 & \textbf{62.88$\pm$4.19}  &87.73$\pm$0.14  &88.27$\pm$0.28 & 82.19\\ 
  mBART\textsubscript{large}& 56.36$\pm$2.77 & 88.19$\pm$0.19 & 86.83$\pm$0.32 & 62.30$\pm$1.24 & 79.24$\pm$0.61 &84.96$\pm$0.31 & 76.31 \\ 
  
  \multicolumn{8}{c}{\textbf{Code-Switching to Code-Switching(CS2CS)}} \\ 
  \textbf{Model} & \textbf{COLA} & \textbf{SST-2} & \textbf{MRPC} & \textbf{RTE} & \textbf{STS-B} & \textbf{QNLI} & \textbf{Avg.} \\ 
  \hline
  mBERT\textsubscript{base}& \textbf{72.48$\pm$2.04} & \textbf{89.83$\pm$1.04} & \textbf{80.80$\pm$1.85} & \textbf{57.31$\pm$4.38} & \textbf{81.77$\pm$1.28} & \textbf{83.91$\pm$0.32} & \textbf{77.68} \\ 
  XLM-R\textsubscript{base}& 72.07$\pm$0.00 & \textbf{91.29$\pm$0.44} & \textbf{85.52$\pm$1.52} & 53.57$\pm$1.39 & \textbf{81.64$\pm$2.36} & \textbf{84.96$\pm$0.15} & \textbf{78.18}  \\ 
  XLM-R\textsubscript{large}&74.35$\pm$1.51 & \textbf{93.69$\pm$0.05} & \textbf{88.68$\pm$1.1} & 60.36$\pm$6.74 & \textbf{88.54$\pm$0.34} & \textbf{90.31$\pm$0.15} & \textbf{82.64} \\ 
  mBART\textsubscript{large}& \textbf{74.37$\pm$0.78} & \textbf{92.60$\pm$0.09} & \textbf{86.85$\pm$0.71} & \textbf{62.38$\pm$0.97} & \textbf{85.09$\pm$0.29} & \textbf{88.39$\pm$0.07} & \textbf{81.61} \\ 

\end{tabular}}
\caption{Comparative performance of various multilingual models on Koglish-GLUE across Different Scenarios: Monolingual (EN2EN), English to Code-Switching (EN2CS), and Code-Switching to Code-Switching (CS2CS). \textbf{Best performances in EN2CS and CS2CS are highlighted.} The MRPC uses the F1-score, STS-B uses Spearman’s correlation, and the remaining tasks use accuracy for performance measurement.}
\label{table : eval CS dataset}

\end{center}
\end{table*}

\subsubsection{Training Details}
\vspace{-0.2cm}
We initiated our experiments based on the checkpoints of four pre-trained multilingual models: mBERT\textsubscript{base}~\cite{devlin2018bert}, XLM-R\textsubscript{base}, XLM-R\textsubscript{large}~\cite{conneau2020unsupervised}, and mBART~\cite{liu2020multilingual}.
We utilized the representation of the ``[CLS]'' token as the final sentence embedding to validate the performance. 
More training details can be found in Appendix~\ref{appendix_A1}.

\vspace{-0.4cm}
\subsubsection{Results}
\label{subsec: Dataset Results}
Table~\ref{table : eval CS dataset} shows the results of the experimental outcomes using the Koglish-GLUE dataset. 
We observed that the models trained and evaluated on English-only data (EN2EN) outperformed those trained and evaluated on a mixed English-Korean Code-Switching dataset (CS2CS). 
This disparity stems from EN2EN's monolingual nature and CS2CS's bilingual complexity, which operates within a Korean context with comparatively fewer resources. 
Additionally, larger models such as XLM-R and mBART exhibit a reduced performance discrepancy between EN2EN and CS2CS. This suggests that the models' performance benefits from the increased diversity of training data spanning multiple languages. 
One of our experiment's standout insights is the pronounced efficacy of the CS2CS method compared to EN2CS across almost all model evaluations. 
This outcome underscores the efficacy of using Code-Switching training in the Koglish dataset in contexts where English-Korean Code-Switching is prevalent.

\subsection{Experiments on ConCSE}
\vspace{-0.1cm}
\label{Experiments : ConCSE}

\subsubsection{Setup}
In this experiment, we utilize the Koglish-NLI dataset for training and the Koglish-STS dataset for evaluation. 
The Koglish-NLI dataset contains triplets of monolingual English sentences (hypothesis, entailment, and contradiction) alongside triplets of code-switched (CS) augmented sentences (CS-hypothesis, CS-entailment, and CS-contradiction).
The Koglish-STS dataset consists of pairs of original sentences (sentence0 and sentence1) and their CS counterparts (CS-sentence0 and CS-sentence1).
During the training phase, we leverage SimCSE~\cite{gao2021simcse} to train the sentence encoder $\mathcal{M}{\phi}$ using CS-augmented sentence triplets.
Moreover, ConCSE trains $\mathcal{M}{\phi}$ on both triplets of original English sentences and CS-augmented sentences, promoting learning in a CS scenario. 
We evaluated both SimCSE and ConCSE using the CS sentence pairs from Koglish-STS.
We adopt Spearman’s correlation as the primary metric for this assessment.

\begin{table*}[!tp] 
\begin{center}
\setlength{\tabcolsep}{2pt}
\renewcommand{\arraystretch}{1.7}
\resizebox{\columnwidth}{!}{
\begin{tabular}{l|c c c c c c c|c} %

  \textbf{Model}& \textbf{STS-B} & \textbf{STS12} & \textbf{STS13} & \textbf{STS14} & \textbf{STS15} & \textbf{STS16} & \textbf{SICK} & \textbf{Avg.} \\ 
  \hline

  SimCSE-mBERT\textsubscript{base} & 77.68$\pm$0.17 & 63.43$\pm$0.22 & \textbf{70.93$\pm$0.33} & 68.59$\pm$0.32 & 79.07$\pm$0.28 & 72.52$\pm$0.28 & 75.91$\pm$0.08 & 72.59\\
  SimCSE-XLM-R\textsubscript{base} & 78.65$\pm$0.30 & 67.49$\pm$0.60 & 75.55$\pm$0.24 & 71.40$\pm$0.18 & 80.03$\pm$0.38 & 76.65$\pm$0.34 & 77.22$\pm$0.20 & 75.29 \\
  SimCSE-XLM-R\textsubscript{large} & 82.43$\pm$0.12 & 71.02$\pm$0.21 & 82.28$\pm$0.32 & 76.19$\pm$0.23 & 83.11$\pm$0.23 & 79.52$\pm$0.25 & \textbf{79.05$\pm$0.20} & 79.09 \\
  *ConCSE-mBERT\textsubscript{base} & \textbf{79.95$\pm$0.24} & \textbf{68.29$\pm$0.21} & 70.52$\pm$1.12 & \textbf{71.24$\pm$0.32} & \textbf{80.40$\pm$0.27} & \textbf{73.13$\pm$0.52} & \textbf{77.01$\pm$0.22} & \textbf{74.36}\\
  *ConCSE-XLM-R\textsubscript{base} & \textbf{79.93$\pm$0.26} & \textbf{71.27$\pm$0.43} & \textbf{75.56$\pm$0.63} & \textbf{74.23$\pm$0.28} & \textbf{80.94$\pm$0.31} & 76.17$\pm$0.22 & \textbf{78.08$\pm$0.16} & \textbf{76.60} \\
  *ConCSE-XLM-R\textsubscript{large} & \textbf{82.85$\pm$0.11} & \textbf{75.00$\pm$0.34} & \textbf{82.72$\pm$0.23} & \textbf{77.80$\pm$0.27} & \textbf{84.12$\pm$0.23} & 79.43$\pm$0.38 & \textbf{78.91$\pm$0.30} & \textbf{80.12}\\
  \hline
  \textbf{p-value(T-test)}& \textbf{STS-B} & \textbf{STS12} & \textbf{STS13} & \textbf{STS14} & \textbf{STS15} & \textbf{STS16} & \textbf{SICK} \\ 
  v.s SimCSE-mBERT\textsubscript{base} & $3.1 \times 10^{-7}$ &  $9.8 \times 10^{-10}$ & 0.508 & $2.8 \times 10^{-6}$ & $1.3 \times 10^{-4}$ & 0.071 & $1.2 \times 10^{-5}$ & \\
  v.s SimCSE-XLM-R\textsubscript{base} & $2.0 \times 10^{-4}$ & $6.8 \times 10^{-6}$ & 0.996 & $1.4 \times 10^{-7}$ & $4.9 \times 10^{-5}$ & 0.046 & $1.4 \times 10^{-5}$ & \\
  v.s SimCSE-XLM-R\textsubscript{large} & $8.7 \times 10^{-5}$ & $4.0 \times 10^{-8}$ & 0.056 & $1.7 \times 10^{-5}$ & $2.0 \times 10^{-5}$ & 0.688 & 0.473 & \\
\end{tabular}
}
\caption{Performance comparison of SimCSE and ConCSE on various Koglish-STS tasks: Performance is measured in terms of Spearman’s correlation in “all” settings. \textbf{Bold values highlight the top performance in each task.} “v.s” in the table is the result of the T-test between SimCSE and ConCSE.}
\label{table: ConCSE}
\end{center}
\end{table*}
\vspace{-0.2cm}

\subsubsection{Training Details}
In our experiments, we initialize our sentence encoder $\mathcal{M}_{\phi}$ using pre-trained mBERT~\cite{devlin2018bert} or XLM-R~\cite{conneau2020unsupervised}, and we use ``[CLS]'' as $\mathcal{M}_{\phi}$ final representation.
We adopt SimCSE~\cite{gao2021simcse} as our baseline model during the implementation phase.
Furthermore, as ConCSE had to handle a larger volume of sentences compared to SimCSE~\cite{gao2021simcse}, we only adjusted the batch size. The rest of the experimental settings were maintained identically to SimCSE. To ensure the accuracy and reliability of our results, we conducted experiments using five different random seeds and recorded the corresponding T-test results.
More details on training can be found in Appendix~\ref{appendix_A2}.

\vspace{-0.5cm}
\subsubsection{Results}
The performance of SimCSE~\cite{gao2021simcse} and ConCSE is summarized in Table~\ref{table: ConCSE}.
In the CS scenario, our proposed ConCSE significantly outperforms SimCSE because it helps implicitly align the representations across languages in CS situations.
Specifically, ConCSE-mBERT\textsubscript{base} outperforms SimCSE-mBERT\textsubscript{base} by improving the average Spearman’s correlation score from 72.59\% to 74.36\%, which significantly outperforms.
We also note consistent performance enhancements with ConCSE-XLM-R\textsubscript{base} and ConCSE-XLM-R\textsubscript{large} models. 
These improvements across all ConCSE backbone models are instrumental in augmenting the comprehension of CS contexts, demonstrating their significant impact. 
Furthermore, The results demonstrate that our ConCSE can scale to multiple datasets and more languages in CS scenarios.

\vspace{-0.3cm}
\subsection{Ablation studies}
In this section, we conduct a comprehensive set of ablation studies to substantiate our ConCSE architecture. Particularly, we evaluated the effects of the combination of the loss function and the effects of temperature, triplet loss, and margin on training by testing the ConCSE-mBERT\textsubscript{base} on the Koglish-STS-B task. 

\subsubsection{Ablation Studies of Loss Functions}
\begin{table}[t]
\centering
\renewcommand{\arraystretch}{1.1}
    \begin{tabular}{@{}c|ccc|c@{}}
    \toprule
    \textbf{-}& \textbf{\quad\;$\mathcal{L}_{CS}^{Con}$\quad\;} & \textbf{\quad\;$\mathcal{L}_{CS}^{Tri}$\quad\;} & \textbf{\quad\;$\mathcal{L}^{Sim}_{neg}$\quad\;} & Koglish-STS-B \\ 
    \midrule
    v1 & \ding{51} & \ding{55} & \ding{55} & 79.63 \\
    v2 & \ding{55} & \ding{51} & \ding{55} & 71.68 \\
    v3 & \ding{51} & \ding{55} & \ding{51} & 79.86 \\
    v4 & \ding{55} & \ding{51} & \ding{51} & 73.53 \\
    v5 & \ding{51} & \ding{51} & \ding{55} & 79.87 \\
    \midrule
    \textbf{ConCSE(v6)} & \ding{51} & \ding{51} & \ding{51} & \textbf{80.23} \\
    SimCSE & \ding{55} & \ding{55} & \ding{55} & 77.45 \\
    \bottomrule
    \end{tabular}
\caption{The Effect of Loss Function Variants on Koglish-STS-B task(Spearman’s correlation, “all” setting).}
\label{table5}
\end{table}
We conducted an ablation study for three types of loss functions: (1) Cross Contrastive Loss $(\mathcal{L}_{CS}^{Con})$, (2) Cross Triplet Loss $(\mathcal{L}_{CS}^{Tri})$, and (3) Align Negative Loss $(\mathcal{L}^{Sim}_{neg})$. Since $\mathcal{L}^{Sim}_{neg}$ cannot be used alone, we analyzed its impact by adding or removing it in different scenarios. Without using all three loss functions, contrastive loss alone results in SimCSE. The performance is recorded in Table~\ref{table5}.
Our findings reveal that only $\mathcal{L}_{CS}^{Con}$ results in a significant performance gain of 2.2\% over the baseline SimCSE. This demonstrates the necessity of $\mathcal{L}_{CS}^{Con}$ in code-switched scenarios.
Also, when comparing v1 and v2, $\mathcal{L}_{CS}^{Tri}$ alone showed a performance decrease (7.9\%) compared to v1. However, when comparing v1 and v5, $\mathcal{L}_{CS}^{Con}$ and $\mathcal{L}_{CS}^{Tri}$ improve performance (0.2\%). 
Similarly, when comparing v1 to v3, $\mathcal{L}_{CS}^{Con}$ and $\mathcal{L}^{Sim}_{neg}$ together improve performance (0.2\%).
The conclusive evidence from our experiments indicates that the most effective sentence embeddings are produced by integrating all three losses, as implemented in our ConCSE(\textbf{v6}), which outperforms the individual loss components.

\subsubsection{Temperature Scaling} 
Temperature is known to play a crucial role in learning~\cite{chen2020simple,gao2021simcse}. To verify this, we conducted a separate ablation study.
Based on our experimental results, ConCSE exhibited optimal performance when the temperature $\tau=0.05$. This value aligns with the temperature used in the learning of SimCSE. Detailed results are in Table~\ref{ablation: Temp scaling}.

\begin{table}[!h]
\centering
    \begin{tabular}{l|cccccc}
    \toprule
    Temperature($\tau$) & \quad 0.001 \quad & \quad 0.01 \quad & \quad \textbf{0.05} \quad & \quad 0.1 \quad & \quad 1 \quad \\
    \midrule
    Koglish-STS-B & \; 76.5
 & \; 77.9 & \; \textbf{80.2} & \quad 78.6
 & \quad 69.8 \\
    \bottomrule
    \end{tabular}
\caption{Temperature scaling for ConCSE across Koglish-STS-B task, evaluated using Spearman’s correlation in the “all” setting.}
\label{ablation: Temp scaling}
\end{table}

\subsubsection{Triplet Loss Scaling} 
During the training of ConCSE, we observed the best performance when using the weight factor of triplet loss $\lambda = 1.2$ . The results are shown in Table~\ref{ablation: Triplet scaling}.

\begin{table}[!h]
\centering
    \begin{tabular}{l|ccccccccc}
    \toprule
    Triplet($\lambda$) & \quad N/A \quad & \quad 1 \quad& \quad \textbf{1.2} \quad & \quad 1.4 \quad & \quad 1.5 \quad & \quad 2 \quad\\
    \midrule
    Koglish-STS-B & \; 79.8 & \quad 80.0 & \quad \textbf{80.2} & \quad 79.8 & \quad 79.9 & \quad 79.8 \\
    \bottomrule
    \end{tabular}
\caption{Performance of ConCSE with different scaling factors for the triplet loss on STS-B(Koglish) tasks, assessed using Spearman’s correlation in the “all” setting.}
\label{ablation: Triplet scaling}
\end{table}

\vspace{-1cm}
\subsubsection{Margin Scaling}
The performance of ConCSE was optimal when using a margin of $\alpha=1$ for the triplet loss, as shown in Table~\ref{ablation:margin scaling}.

\begin{table}[!h]
\centering
\begin{tabular}{l|ccccccccc}
    \toprule
    Margin($\alpha$) & \quad 0.5 \quad & \quad \textbf{1} \quad & \quad 1.2 \quad & \quad 1.4 \quad & \quad 1.5 \quad & \quad 2 \quad \\
    \midrule
    Koglish-STS-B & \quad 79.7 & \quad \textbf{80.2} & \quad 80.0 & \quad 79.9 & \quad 79.9 & \quad 79.7 \\
    \bottomrule
    \end{tabular}
\caption{Performance of ConCSE with different margin values for the triplet loss on Koglish-STS-B task (Spearman’s correlation, “all” setting).}
\label{ablation:margin scaling}
\end{table}

\section{Conclusion}
\vspace{-0.3cm}
In this work, we first introduced the novel Koglish dataset, focusing on code-switching (CS) between English-Korean and Korean-English. This Koglish dataset marks an initial pioneering attempt, and exhaustive evaluations have highlighted the critical need for such a resource. Second, we propose a method to learn universal code-switched sentence embeddings using this newly constructed Koglish dataset. Surprisingly, Through extensive testing, ConCSE surpassed other leading sentence embedding techniques in Koglish-STS tasks. Nevertheless, our study has certain constraints: Although less frequent, grammatical elements other than nouns or noun phrases can also be CS in English-Korean CS situations. In our future work, we aim to develop a more comprehensive CS dataset encompassing all grammatical elements. We are optimistic that our contributions will spur further research and progress in the understanding and application of low-resource CS data.
\vspace{-0.2cm}

%
%
%
\bibliographystyle{splncs04}
\bibliography{refernece}

\clearpage
\appendix
\section{appendix}

\subsection{Training Details on Koglish dataset}
\label{appendix_A1}
For each experiment, we used a pair of NVIDIA RTX A6000 GPUs. Furthermore, we trained with 5 epochs, used batch size $\in \{32,64,128\}$ and learning rate $\in \{5\text{e}-5,4\text{e}-5,3\text{e}-5,1\text{e}-5\}$ with grid search, and evaluated the model using five randomized seeds to ensure the experiment's reproducibility. Hyperparameter settings can be found in Table~\ref{table: Koglish Dataset Training Details}.


\begin{table}[H]
\begin{center}
    \begin{tabular}{l|c c}
      \toprule
      \textbf{Model. Param} & Batch Size & Learning Rate \\
      \midrule
      mBERT\textsubscript{base} & 128 & 5e-5 \\
      XLM-R\textsubscript{base} & 128 & 5e-5 \\
      XLM-R\textsubscript{large} & 64 & 1e-5 \\
      mBART\textsubscript{large} & 64 & 1e-5 \\
      \bottomrule
    \end{tabular}
\caption{Hyperparameter specifications for batch size and learning rate during training on the Koglish-GLUE.}
\label{table: Koglish Dataset Training Details}
\end{center}
\end{table}

\subsection{Training Details of ConCSE}
\label{appendix_A2}
For each experiment, we utilized a pair of NVIDIA RTX A6000 GPUs.
Furthermore, we trained with 5 epochs, used batch size $\in \{32,64,128\}$ and learning rate $\in \{5\text{e}-5,4\text{e}-5,3\text{e}-5,1\text{e}-5\}$ with grid search on the validation set of Koglish-STS-B.
Hyperparameter settings can be found in Table~\ref{table: SimCSE and ConCSE Training details}.
Referring to the findings from previous research by~\cite{chen2020simple}, while contrastive learning typically benefits from larger batch sizes, we observed that ConCSE yielded effective results even with smaller batch sizes. 

\begin{table}[h]
\begin{center}
\begin{tabular}{l|c c c}
  \toprule
  \textbf{Param. Model} & \textbf{mBERT\textsubscript{base}} & \textbf{XLM-R\textsubscript{base}} & \textbf{XLM-R\textsubscript{large}} \\
  \midrule
  Batch Size (SimCSE) & 512 & 512 & 128 \\
  Learning Rate (SimCSE) & 5e-5 & 5e-5 & 1e-5 \\
  \midrule
  Batch Size (ConCSE) & 128 & 128 & 48 \\
  Learning Rate (ConCSE) & 5e-5 & 5e-5 & 1e-5 \\
  \bottomrule
\end{tabular}
\caption{Specification about the batch size and learning rates for SimCSE and ConCSE.}
\label{table: SimCSE and ConCSE Training details}
\end{center}
\end{table}
%




\end{document}